# ELECTRE TREE – A MACHINE LEARNING APPROACH TO INFER ELECTRE TRI-B PARAMETERS

Gabriela Montenegro de Barros
gabimontenegro@eq.ufrj.br
Federal Fluminense University

Valdecy Pereira
valdecypereira@id.uff.br
Federal Fluminense University

**Abstract:**

*Purpose*: This paper presents an algorithm that can elicitate (infer) all or any combination of ELECTRE Tri-B parameters. For example, a decision-maker can maintain the values for indifference, preference, and veto thresholds, and our algorithm can find the criteria weights, reference profiles, and the lambda cutting level. Our approach is inspired by a Machine Learning ensemble technique, the Random Forest, and for that, we named our approach as ELECTRE Tree algorithm. *Methodology*: First, we generate a set of ELECTRE Tri-B models, where each model solves a random sample of criteria and alternatives. Each sample is made with replacement, having at least two criteria and between 10% to 25% of alternatives. Each model has its parameters optimized by a genetic algorithm that can use an ordered cluster or an assignment example as a reference to the optimization. Finally, after the optimization phase, two procedures can be performed, the first one will merge all models, finding in this way the elicitated parameters, and in the second procedure each alternative is classified (voted) by each separated model, and the majority vote decides the final class. *Findings*: We have noted that concerning the voting procedure, non-linear decision boundaries are generated, and they can be suitable in analyzing problems with the same nature. In contrast, the merged model generates linear decision boundaries. *Originality*: The elicitation of ELECTRE Tri-B parameters is made by an ensemble technique that is composed of a set of multicriteria models that are engaged in generating robust solutions.

# 1- Introduction

Decision Making can turn out to be a complex problem, especially if the consequences of a poor decision are severe and significant (Govindan & Jepsen, 2016). Nonetheless, Wang et al. (2009) affirm that to reach a reasonable decision, the discipline Multiple Criteria Decision Analysis (MCDA) can offer a series of techniques that can evaluate many courses of action (alternatives) under conflicting points of view (criteria).

According to Mendoza and Martins (2006), numerous MCDA methods have been proposed in the literature, like Analytic Hierarchy Process (AHP), Multi-attribute Utility Theory (MAUT), Simple Additive Weighting (SAW), Elimination and Choice Expressing Reality (ELECTRE), Technique for Order of Preference by Similarity to Ideal Solution (TOPSIS), Preference Ranking and Organization Method for Enrichment Evaluation (PROMETHEE) and many others creations, variations, and extensions.

In this study, we choose to focus on one method of the ELECTRE family, the ELECTRE Tri-B (formerly know as ELECTRE Tri), created by Yu (1992) and further developed by Roy.and Bouyssou (1993). This method can divide a set of alternatives into ordered classes, for example, Class A, Class B, Class C, where Class A > Class B > Class C. Naturally, we can affirm that the alternatives allocated into Class A are better than the ones allocated in Classes B and C. Many types of essential decision-making problems can be solved by this method, for example, industrial risk classification, classification of products in storage, classification of a group of suppliers, and others.

Fernández et al. (2019) indicate that many other methods can be applied to this problem. However, from the ones that exploit outranking relations, the ELCETRE Tri- B method is, perhaps, the most widely used approach.

This type of problem to be solved by the ELECTRE Tri-B method needs to set up several parameters: indifference, preference, and veto thresholds, criteria weights, number of reference profiles, reference profiles values, and the lambda cutting level The time-consuming task of deciding these parameters may drive the decision-maker away from the method or make him/her chose impulsive values without any polishment. In this context, we propose an algorithm that can elicitate, excluding the number of profiles, all or any combination of ELECTRE Tri-B parameters.

However, we emphasize that although tempting, the elicitation of all parameters is not indicated. Some intuition and knowledge that the decision-maker has about the problem must be preserved. The decision-maker or decision-makers should decide the set of parameters that they are willing to preserve, corroborating with the main idea of Roy et al. (2014) that discussed the importance of the thresholds as tools used by decision-makers to model imperfect knowledge.

Inspired by a Machine Learning ensemble technique, specifically the Random Forest, a method that was proposed by Ho (1995), our algorithm – named ELECTRE Tree – also starts with a divide and conquer strategy. Generating samples with replacement of alternatives and criteria, feeding each one of these samples to its respective ELECTRE Tri-B model, and optimizing the unknown parameters via a genetic algorithm. The optimization is based on the intuition that similar alternatives have a higher chance of being in the same class. Ordered clusters are generated adjusting the K-means++ algorithm, and the optimal parameters are found when the ELECTRE Tri-B solution has the maximum possible alignment with the clusterization solution. As a viable alternative, assignments examples can be used in lieu of the clusterization procedure. Finally, after the optimization phase, each alternative is classified by each model, and the majority vote decides the class that an alternative belongs.

The article is divided into five sections, besides the introduction, as follows. Section 2 presents a literature review considering the elicitation of parameters problem in the ELECTRE Tri B method. In sectio.n 3, we describe our algorithm in detail. Section 4 presents the case study and the obtained results, and finally, in section 5, we make our final considerations.

## 2- Literature Review

The literature shows various methodologies to deal with the parameters elicitation problem. Some studies that, try, mathematically to find the parameters' values. Other studies combine the ELECTRE Tri-B method with different techniques to have initial values for some parameters. For example, Jerônimo and Medeiros (2014) uses the results of a SERVPERF (Service Performance) questionnaire to generate weights to the ELECTRE Tri-B method. Even though these are valid strategies to deal with the parameters elicitation problem, we chose to focus on the literature review, in studies that are mathematically developed to find these parameters.

Mousseau and Slowinski (1998) argue how the direct elicitation of the ELECTRE Tri-B parameters is difficult, and the authors propose a methodology to elicit them indirectly. Their approach involves an interactive approach that infers the parameters of an ELECTRE Tri-B model from assignment examples, requiring less cognitive effort from the decision-maker. The intuition behind this approach is that the decision-maker prefers to give some assignment examples rather than to specify the values of parameters directly. However, the veto threshold is not considered because of computational complexity. In the last step, a formulated non-linear optimization program finds the final form of the ELECTRE Tri-B model.

Later Mousseau et al. (2001) extended the work of Mousseau and Slowinski (1998) by considering the subproblem of the determination of the weights only. This subproblem led to the formulation of a linear optimization program with stable results. Also, they show that the number of assignment examples equal to double the number of criteria is sufficient to get to a solution. Dias and Mousseau (2002), in the same way, developed a linear optimization program to infer the veto thresholds for the ELECTRE Tri-B and ELECTRE III models. Ngo The and Mousseau (2002) complements the works of Mousseau et al. (2001) and Dias and Mousseau (2002) by creating a couple of models that can, in an interactive approach, elicitate the reference profiles of ELECTRE Tri-B model.

Mousseau et al. (2000) developed a software to reduce the cognitive effort required from the decision-maker in the phase of calibration of the ELECTRE Tri-B model. The parameters, weights of criteria and the lambda cutting level (pessimistic assignment procedure only), are obtained from an optimization problem derived from assignment examples supplied by the decision-maker

Doumpos et al. (2009) proposed an evolutionary approach to elicitate appropriate ELECTRE Tri-B parameters. Their method also allows the addition of constraints about these parameters; thus, the decision-maker is free to use his/her previous obtained preference information, such as weights, reference profiles, and thresholds. However, to their evolutionary approach work, a set of assignment examples are needed.

Cailloux et al. (2012) suggested an elicitation procedure to infer two ELECTRE Tri-B parameters, the reference profiles, and, optionally, veto thresholds. The elicitation takes into account assignment examples provided by multiple decision-makers to formulate a linear optimization program. According to their results, they perform well on datasets corresponding to real-world decision problems. Nevertheless, they conclude that datasets involving more than eight criteria are still challenging to solve.

Leone and Minnetti (2013) proposed a procedure composed of two phases to infer the ELECTRE Tri-B parameters. In the first phase, a linear optimization program is dedicated to elicitate profiles and threshold values. In the second phase, a non-linear optimization program is dedicated to elicitate weights and lambda cutting level. The main advantage of the two phases approach is that the decision-maker keeps better control of the parameters and can fix some parameters; for example, the weights.

Zheng et al. (2014) proposed a linear optimization program procedure to elicitate criteria weights and the lambda cutting level for the optimistic rule of the ELECTRE Tri-B method. The authors also provided an algorithm that computes robust alternatives' assignments from assignment examples. Additionally, their experiments pointed out that to accurately generate weights, a significant number of assignment examples are required.

Fernández et al. (2019) use a genetic algorithm to solve a non-linear optimization program that adopts assignment examples, to infer the ELECTRE Tri-nB (a variant of the original method) veto threshold, weights, lambda cutting level and reference profile parameters.

Ramezanian (2019) finds the reference profiles in the *posteriori* ELECTRE Tri-B method. This method needs, in the first step, that a decision-maker assigns the alternatives to predefined classes. Then these assignments are used in a non-linear optimization program; however, due to the complexity of the optimization model, the Particle Swarm Optimization (PSO) metaheuristic is employed.

Leroy et al. (2011) developed a simplified variant of the ELECTRE Tri-B method that they called MR-Sort. To elicit the values of the reference profiles and weights, they consider a linear programming model that learns from a set of known assignment examples. Sobrie et al. (2019) studied this simplified variant of the ELECTRE Tri-B method, and they have described a metaheuristic developed to learn its parameters. Their metaheuristic improved the computing time and memory space resources when compared with the alternative linear programming model.

The literature review, although not exhaustive, was able to indicate some tactics used to solve the parameters elicitation problem. These tactics used isolated or in combination, involved: targeting only a set of parameters, using assignment examples generated by experts, using assignment examples generated by specialized algorithms or software, formulating linear optimization programming, formulating non-linear optimization programming, applying metaheuristics to solve the mathematical models, choosing only the pessimistic rule case, choosing only the optimistic rule case, ignoring veto thresholds, dividing the problem into subproblems, and, simplifying the original ELECTRE Tri-B method.

Our proposed model relates to the other ones in the literature in the sense that it also applies assignment examples - generated by the decision-maker or generated by the clusterization task -, and a metaheuristic procedure to solve a set of complex subproblems. Nonetheless, our model differs from the other ones in the literature in the sense that it uses a specific Machine Learning technique as a way to create a set of different models that are employed altogether to find the final solution.

**3 – Methodology**

The ELECTRE Tree algorithm is a composition of the following techniques:1) ELECTRE Tri-B: where all or a set of parameters can be elicitated using the optimistic or pessimistic rules; 2) A clusterization algorithm or a set of assignment examples: if clusters are used instead of assignment examples, then the K-means++ algorithm, due to its simplicity and computation time performance; 3) A metaheuristic: where the genetic algorithm was used, but other ones can be used instead.

Before we delve into the ELECTRE Tree steps, each technique, ELECTRE Tri-B, K-means++, and Genetic Algorithm are, respectively, explained in the following subsections of Section 3.

### 3.1 –. ELECTRE Tri-B Method

The ELECTRE Tri-B method was developed to solve ordered classification problems by comparing alternatives with reference profiles, which form the boundaries of each established class (category). Considering that $G = \{g_1, g_2, ..., g_j\}$ is a set containing $j$ criteria and each one with a weight (level of importance) $w_j$, let $X = \{x_1, x_2, x_3, ..\}$ be a vector in $R^j$ that represents the evaluations of a generic alternative $x$ under each criterion in $G$, and finally, considering that $B = \{b_1, b_2, ..., b_n\}$ is a set of $n + 1$ reference profiles where $b_{h-1}$ and $b_h$ are, respectively, the lower and the upper limits to the $h^{th}$ class.

For each $h$ class and each $j$ criterion, $g_j(b_h)$, represents the evaluation of the upper limit of the $h^{th}$ class by the $j^{th}$ criterion. For each alternative $x_i$ and each $j$ criterion, $g_j(x_i)$, represents the evaluation of $i^{th}$ alternative for the $j^{th}$ criterion. The outranking depends on the absolute value of the differences $g_j(x_i)$-$g_j(b_h)$, being higher than predetermined indifference, preference, and veto thresholds $q_j$, $p_j$ and $v_j$, where $v_j \geq p_j \geq q_j$. Then the following steps are performed to get the outranking relations.

a) Compute the partial concordance degree $c_j(x_i, b_h)$ and $c_j(b_h, x_i)$ (equations $1a$ and $1b$).

$$c_j(x_i, b_h) = \begin{cases} 0, & \text{if } g_j(b_h) - g_j(x_i) \geq p_j \\ 1, & \text{if } g_j(b_h) - g_j(x_i) < q_j \\ \dfrac{p_j - g_j(b_q) + g_j(x_i)}{p_j - q_j}, & \text{if } q_j \leq g_j(b_h) - g_j(x_i) < p_j \end{cases} \quad (1a)$$

$$c_j(b_h, x_i) = \begin{cases} 0, & \text{if } g_j(x_i) - g_j(b_h) \geq p_j \\ 1, & \text{if } g_j(x_i) - g_j(b_h) < q_j \\ \dfrac{p_j - g_j(x_i) + g_j(b_q)}{p_j - q_j}, & \text{if } q_j \leq g_j(x_i) - g_j(b_h) < p_j \end{cases} \quad (1b)$$

b) Compute the global concordance degree $C(x_i, b_h)$ and $C(b_h, x_i)$ (equations $2a$ and $2b$).

$$C(x_i, b_h) = \frac{\sum_{j=1}^{n} w_j c_j(x_i, b_h)}{\sum_{j=1}^{n} w_j} \quad (2a)$$

$$C(b_h, x_i) = \frac{\sum_{j=1}^{n} w_j c_j(b_h, x_i)}{\sum_{j=1}^{n} w_j} \quad (2b)$$

c) Compute the partial discordance degree $D_j(x_i, b_h)$ and $D_j(b_h, x_i)$ (equations $3a$ and $3b$).

$$D_j(x_i, b_h) = \begin{cases} 0, & \text{if } g_j(b_h) - g_j(x_i) < p_j \\ 1, & \text{if } g_j(b_h) - g_j(x_i) \geq v_j \\ \dfrac{-p_j - g_j(x_i) + g_j}{v_j - p_j}, & \text{if } p_j \leq g_j(b_h) - g_j(x_i) < v_j \end{cases} \quad (3a)$$

$$D_j(b_h, x_i) = \begin{cases} 0, & \text{if } g_j(x_i) - g_j(b_h) < p_j \\ 1, & \text{if } g_j(x_i) - g_j(b_h) \geq v_j \\ \dfrac{-p_j - g_j(b_q) + g_j}{v_j - p_j}, & \text{if } p_j \leq g_j(x_i) - g_j(b_h) < v_j \end{cases} \quad (3b)$$

d) Compute the credibility degree $\sigma(x_i, b_h)$ and $\sigma(b_h, x_i)$, which expresses the confidence in the statement "$x_i$ is not worse than $b_h$" (equations $4a$ and $4b$).

$$\sigma(x_i, b_h) = \begin{cases} C(x_i, b_h) \times \prod_{j=1}^{n} \dfrac{1 - D_j(x_i, b_h)}{1 - C(x_i, b_h)}, & \text{if } D_j(x_i, b_h) > C(x_i, b_h) \\ C(x_i, b_h), & \text{Otherwise} \end{cases} \quad (4a)$$

$$\sigma(b_h, x_i) = \begin{cases} C(b_h, x_i) \times \prod_{j=1}^{n} \dfrac{1 - D_j(b_h, x_i)}{1 - C(b_h, x_i)}, & \text{if } D_j(b_h, x_i) > C(b_h, x_i) \\ C(b_h, x_i), & \text{Otherwise} \end{cases} \quad (4b)$$

e) The outranking decision still depends on a last previously determined parameter, the lambda cutting level ($\lambda$) (equation 5):

$$x_i S b_h \text{ if } \sigma(x_i, b_h) \geq \lambda; \quad 0.5 \leq \lambda \leq 1 \quad (5)$$

Finally, the assignment of each alternative to a class can be obtained by two different rules, pessimistic and optimistic rules. In the pessimistic rule, an alternative is compared successively with each $b_h$ reference profile, for $h = n, n-1, \ldots, 0$. When $x_i S b_h$, it is considered that the alternative belongs to the class $C_{h+1}$. In the optimistic rule, an alternative is compared successively with each $b_h$ reference profile, for $h = 1, 2, \ldots, n$. When $b_h S x_i$, it is considered that the alternative belongs to the class $C_h$. In general, both procedures may disagree about the classification of the alternatives, and their classification may not be unique.

**3.2 – K-means and K-means++ Algorithms**

The K-means algorithm was developed by McQueen (1967), and it can group objects into a collection of $k$ clusters based in a distance measure, the number of clusters, $k$, is a parameter. The extension of K-means, named K-means++ was proposed by Arthur and Vassilvitskii (2007) and overcame the K-means' inherent weakness of forming ill-defined clusters. The pseudocode of the K-means++ algorithm is shown in Table 01.

Table 01: K-means++ pseudocode

01. **INPUT** Select the number of clusters $k$, where $K = \{C_1, \cdots, C_k\}$.
02. Define the first cluster centroid $C_1$ as a random observation of the dataset
03. **FOR** every $k$ cluster centroid **DO**
04.     Choose a new centroid for $C_k$ selecting an observation of the dataset with probability $\frac{D(x)^2}{\sum D(x)^2}$, Where $D(x)$ denotes the shortest distance from an observation to the closest centroid already chosen.
05 **OUTPUT** All clusters centroids.

The careful steps taken to calculate clusters centroids avoid the merges of the same, leading to ill-defined group formations. Then, the standard K-means algorithm, described in Table 02, can be executed. Although the initial selection in the algorithm takes extra steps, the k-means algorithm converges very quickly after the centroid definitions (ARTHUR &VASSILVITSKII, 2007).

Table 02: K-means pseudocode

01. **INPUT** $K = \{C_1, \cdots, C_k\}$ clusters centroids
02. **FOR** every instance in the dataset and following the instance order **DO**
03.     Reassign the observation to its closest cluster centroid, $x_i \in C_s$ is moved $C_s$ to $C_t$ if $\|x_i - \bar{x}_t\| \leq \|x_i - \bar{x}_j\| \ \forall j = 1, \cdots, k, j \neq s$
04.     Recalculate centroids for clusters $C_s$ and $C_t$
05. **IF** cluster membership is stabilized **THEN** stop
06.     **ELSE** go to line 02
07. **OUTPUT** Instance assignment

As demonstrated in Table 02, the K-means algorithm the $k$ clusters centroids values allocate each observation in the cluster represented by the nearest centroid reducing in this way the square-error. When the cluster membership of an instance changes, then the centroids of the clusters $C_s$ and $C_t$, needs to be recalculated, and the same is required for the square-error. These steps are repeated until convergence; therefore, the square-error cannot be further reduced, and no instance can change its cluster membership (PEÑA et al. 1999).

However, the K-means++ algorithm does not impose a hierarchy between the formed clusters, but this order can be easily obtained by taking into account the monotone nature of the ELECTRE Tri-B problems. It suffices to get the centroids of each cluster, then calculates each centroid distance from the origin and rank in decreasing order, as centroids far from the origin belong to better classes than the ones near the origin. The resulting ordered cluster works as a reference, that's is reasonable enough, to guide the genetic algorithm in the optimization phase.

### 3.3 – Genetic Algorithm

John Henry Holland (February 2, 1929 – August 9, 2015) was the father of the metaheuristic called Genetic Algorithms (GA). Holland popularized this term in his paper published in 1973, entitled "Genetic Algorithms and the Optimal Allocation of Trials". Lucasius and Kateman (1993) affirm that GA is highly suited to tackling complex, large-scale optimization problems in an efficient and efficacious way. Burke and Varley (1997) also pointed out that the main strength of the GA is replicating Darwin's theory of evolution, which obeys the principles of natural selection and survival of the fittest.

Usually, a problem solution is represented by a vector of genes, and each gene corresponds to a value of a variable present in a target function that needs to be optimized. A collection of genes is called a chromosome, and a group of chromosomes is called population. Each chromosome has a fitness indicator that corresponds to the target function value obtained when the genes are plugged in. Therefore the main idea is to reach a population that is dominated by solutions that optimize the target function.

Konak et al. (2006) indicate the steps to execute a GA; first, the population of size $N$ needs to be randomly initialized. Then to create the next population generation, pairs of chromosomes (parents) are combined until the next population reaches the size $N$. Chromosomes with better fitness indicators have a higher chance of reproducing. The combination of a pair of chromosomes, or breeding, uses an operation called crossover to generate better offsprings, in other words, better solutions. Each chromosome also has a chance to mutate at the gene level. This mutation operation introduces random changes that bring genetic diversity into the population, that are vital to escape from local optima.

Of course, this is a general framework to apply GA in any problem, and for this work, we have specific and directions and adaptations involving each step of the GA. As indicated by Pirlot (1996) a metaheuristic adapted to the problem structure performs better.

Concerning the target function, we will use accuracy as a measure to attest to the quality of the solution. The accuracy is calculated by comparing the assignments of the ELECTRE Tri-B model with the aligned solution of the K-means++ assignment, as demonstrated in section 3.2.1. One can note that we can also use predefined assignments or assignments examples, instead of the K-means++. Reasonable solutions have accuracy values as near as possible to 1 (100% accuracy). Although simple, this target function proved to be an essential metric to guide the GA to the convergence of robust parameters

For the breeding phase, the chosen crossover procedure was the Simulated Binary Crossover (SBX) operator. The SBX operator was developed by Deb and Agrawal (1995) to deal with real coded values. Their study showed that the SBX operator was able to perform as good or better than binary-coded GA with the single-point crossover. Moreover, it was found that this operator is indicated to problems that have multiple optimal solutions and problems where the lower and upper bounds of the global optimum are not known a priori. The SBX operator has the following steps:

1) Set a random number $\xi \sim (0,1)$

2) Use $\mu$ and $\eta$ to calculate the beta value $\beta_i$ for each $i$ gene, according to equation 6:

$$\beta_i = \begin{cases} (2\mu)^{\frac{1}{\eta+1}} & if\ \xi < 0.5 \\ \left(\frac{1}{2-2\mu}\right)^{\frac{1}{\eta+1}} & if\ \xi \geq 0.5 \end{cases} \tag{6}$$

3) Generate an offspring using equation 7.

$$\text{offspring}_i = \begin{cases} \dfrac{(1 - \beta_i) \times \text{parent}_1 + (1 + \beta_i) \times \text{parent}_2}{2} & if\ \xi < 0.5 \\ \dfrac{(1 + \beta_i) \times \text{parent}_1 + (1 - \beta_i) \times \text{parent}_2}{2} & if\ \xi \geq 0.5 \end{cases} \quad (7)$$

The offspring may have values that are out of the lower or upper bounds, define by decision-maker or that violates the ELECTRE Tri-B constraints about the thresholds or the reference profile values. To avoid these violations, a clipping operation verifies and corrects the values of each gene. For example, suppose that $q_1$ equals 5 and $p_1$ equals 4. The clipping operation will correct $p_1$ by changing its value to 5, respecting the ELECTRE Tri-B constraint: $v_j \geq p_j \geq q_j$.

Concerning the mutation procedure, the Real-coded Jumping Gene Genetic Algorithm (RJGGA) operator was chosen. The RJGGA operator was developed by Ripon et al. (2007), and the main feature of RJGGA is that it is a simple operation in which a transposition of a gene is induced within the same or another chromosome. The RJGGA operator has the following steps:

1) Set two random numbers $\phi_{ij} \sim (0,1)$ and $\xi_{ij} \sim (0,1)$ for each $j$ gene of every chromosome $i$.

2) For every $\text{gene}_{ij}$, select a low probability (1%, for example) to a mutation occur. Then if $\phi_{ij} \leq 0.01$, mutate the $\text{gene}_{ij}$ using equations 8 and 9.

$$\text{gene}_{ij} = \text{gene}_{ij} + \rho_{ij} \quad (8)$$

$$\rho_{ij} = \begin{cases} (2\xi_{ij})^{\frac{1}{\eta+1}} - 1 & if\ \xi_{ij} < 0.5 \\ 1 - (2 - 2\xi_{ij})^{\frac{1}{\eta+1}} & if\ \xi_{ij} \geq 0.5 \end{cases} \quad (9)$$

Again, after the mutation, the clipping operator needs to verify and correct each gene. As an example, suppose that we want to find the values for the weights ($0 \leq W \leq 1$), the preference threshold $p_j$, the veto threshold, and one reference profile of a problem with two criteria and one reference profile. The decision-maker choose to set the indifference threshold $q_j$ as 0 and the lambda cutting level, $\lambda$, as 0.75.

The problem representation, illustrated in Figure 01, shows that the chromosome vector has the following gene order: weights, indifference threshold, preference threshold, veto threshold, reference profile, and lambda cutting level.

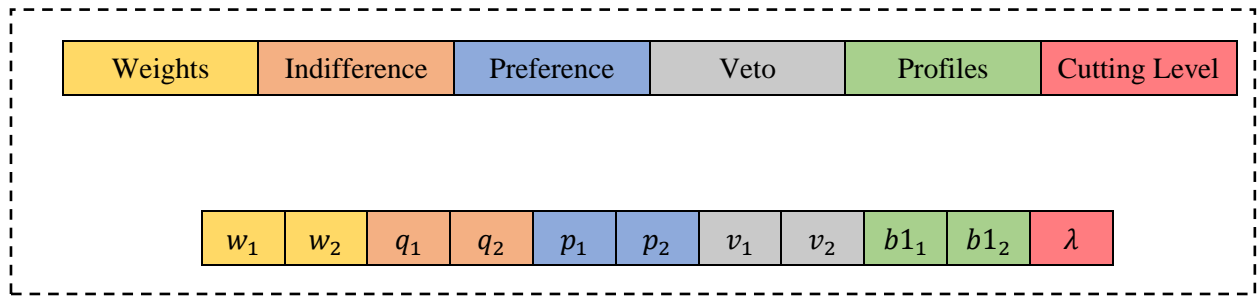

Figure 01: Problem Representation

Moreover, Figure 02 shows the clipping operator limits. The weights can never be lower than 0 or higher than 1, $q_j$ and $\lambda$ can not change their values, and the other parameters can vary, respecting the ELECTRE Tri-B constraints, in other words, only feasible solutions are created.

| $min$ | 0 | 0 | 0 | 0 | $p_1$ | $p_2$ | $v_1$ | $v_2$ | $b1_1$ | $b1_2$ | 0.75 |
| $max$ | 1 | 1 | 0 | 0 | $p_1$ | $p_2$ | $v_1$ | $v_2$ | $b1_1$ | $b1_2$ | 0.75 |

Figure 02: Gene Limits

Optionally an elitist scheme can be adopted, where a number $p$ of fittest individuals of the parent population are preserved in the next generation. However, depending on the problem, if $p$ is too large, it can make the solutions stagnate at a local optimal point. Additionally, our stop criterion is to set a maximum number of generations, but any other stop criterion can be used.

### 3.4. Elicitation Algorithm – ELECTRE Tree

The ELECTRE Tree Algorithm, described in Table 06, with all other techniques described and understood, is straightforward to use.

Table 03 – ELECTRE Tree Algorithm pseudocode

| |
|---|
| 01. **INPUT** Define the ELECTRE Tri-B parameters that need to be inferred; Sample size $S$; Number of models $N$ Assignment Examples or Ordered Cluster, and the GA parameters. The criteria need to be from maximization type. |
| 02. **FOR** $i = 0$ to $N$ **DO** |
| 03.     Sample with replacement $S$ observations from the population |
| 04.     Sample with replacement $k$ criterion, where $k \geq 2$ |
| 05.     Run the ELECTRE Tri-B model with the sampled criteria and alternatives |
| 06.     **IF** Assignments Examples are given: |
| 07.         Use the assignment examples as reference |
| 08.     **ELSE**: |
| 09.         Use the ordered clusters as reference |
| 10.     Generate the ELECTRE Tri-B unknown parameters using the GA that uses as loss function the accuracy between the ELECTRE Tri-B and the reference |
| 11.     Include the model with the converged parameters in the solution list |
| 12. **FOR** each $j$ alternative **DO** |
| 13.     **FOR** $i = 0$ to $N$ **DO** |
| 14.         Classify the alternatives using model $i$ |
| 15.         Count the class votes for the alternative $j$ |
| 16. **RETURN** votes, and elicited parameters |

Line 01 of Table 03, indicates that we must define which parameters we need to elicitate, the sample alternatives sample size, usually, must be between 10% and 25%, the number of models, generally between 200 to 1000 models is enough to reach a solution. The recommended initial GA parameters are: 25 to 250 generations, 10 to 50 chromosomes as the population size, one elite member, the operator parameters $\mu$ equal 2 and $\eta$ equal 1, and finally, the mutation rate between 1% and 5%. Depending on the problem, the decision-maker may calibrate these initial values, as he/she wishes.

From line 02 to line 11, we randomly select a total of $S$ alternatives and $k$ criteria. The minimum size of $k$ is two. Then we find the model parameters by optimizing the accuracy that compares the ELECTRE Tri-B solution and the clusters/assignment examples. We repeat this process $N$ times, obtaining this way a solution list with all optimized models. Finally, from line 12 to line 16, the alternatives final classes are decided by majority vote or by the model with the elicitated parameters.

## 4 – Numerical Examples

To illustrate our methodology three examples are described, the first one in Table 04, is a synthetic dataset elaborated to explain step by step our algorithm, this dataset possesses 64 alternatives and two criteria, where both criteria are from the maximization type. Also, it is separated in 4 classes A, B, C, and D, where Class A > Class B > Class C > Class D. No assignment examples are given, the pessimistic rule is used, and we must all parameters need to be elicitated.

Table 04: Dataset 1

| Alt. | $g1$ | $g2$ | Alt. | $g1$ | $g2$ | Alt. | $g1$ | $g2$ | Alt. | $g1$ | $g2$ |
|---|---|---|---|---|---|---|---|---|---|---|---|
| $x1$ | 1 | 1 | $x17$ | 16 | 8 | $x33$ | 23 | 8 | $x49$ | 23 | 15 |
| $x2$ | 1 | 2 | $x18$ | 16 | 9 | $x34$ | 23 | 9 | $x50$ | 23 | 16 |
| $x3$ | 1 | 3 | $x19$ | 16 | 10 | $x35$ | 23 | 10 | $x51$ | 23 | 17 |
| $x4$ | 1 | 4 | $x20$ | 16 | 11 | $x36$ | 23 | 11 | $x52$ | 23 | 18 |
| $x5$ | 2 | 1 | $x21$ | 17 | 8 | $x37$ | 24 | 8 | $x53$ | 24 | 15 |
| $x6$ | 2 | 2 | $x22$ | 17 | 9 | $x38$ | 24 | 9 | $x54$ | 24 | 16 |
| $x7$ | 2 | 3 | $x23$ | 17 | 10 | $x39$ | 24 | 10 | $x55$ | 24 | 17 |
| $x8$ | 2 | 4 | $x24$ | 17 | 11 | $x40$ | 24 | 11 | $x56$ | 24 | 18 |
| $x9$ | 3 | 1 | $x25$ | 18 | 8 | $x41$ | 25 | 8 | $x57$ | 25 | 15 |
| $x10$ | 3 | 2 | $x26$ | 18 | 9 | $x42$ | 25 | 9 | $x58$ | 25 | 16 |
| $x11$ | 3 | 3 | $x27$ | 18 | 10 | $x43$ | 25 | 10 | $x59$ | 25 | 17 |
| $x12$ | 3 | 4 | $x28$ | 18 | 11 | $x44$ | 25 | 11 | $x60$ | 25 | 18 |
| $x13$ | 4 | 1 | $x29$ | 19 | 8 | $x45$ | 26 | 8 | $x61$ | 26 | 15 |
| $x14$ | 4 | 2 | $x30$ | 19 | 9 | $x46$ | 26 | 9 | $x62$ | 26 | 16 |
| $x15$ | 4 | 3 | $x31$ | 19 | 10 | $x47$ | 26 | 10 | $x63$ | 26 | 17 |
| $x16$ | 4 | 4 | $x32$ | 19 | 11 | $x48$ | 26 | 11 | $x64$ | 26 | 18 |

Figure 03 shows the plot of the dataset points, $g_1$ as the abscissa and $g_2$ as the ordinate. It can be noticed that the original data structure arrangement implies that, the alternatives from $x1$ to $x16$ belong to class D, the alternatives from $x17$ to $x32$ belong to Class C, the alternatives from $x33$ to $x48$ belong to Class B, and finally, the alternatives from $x49$ to $x64$ belongs to Class A.

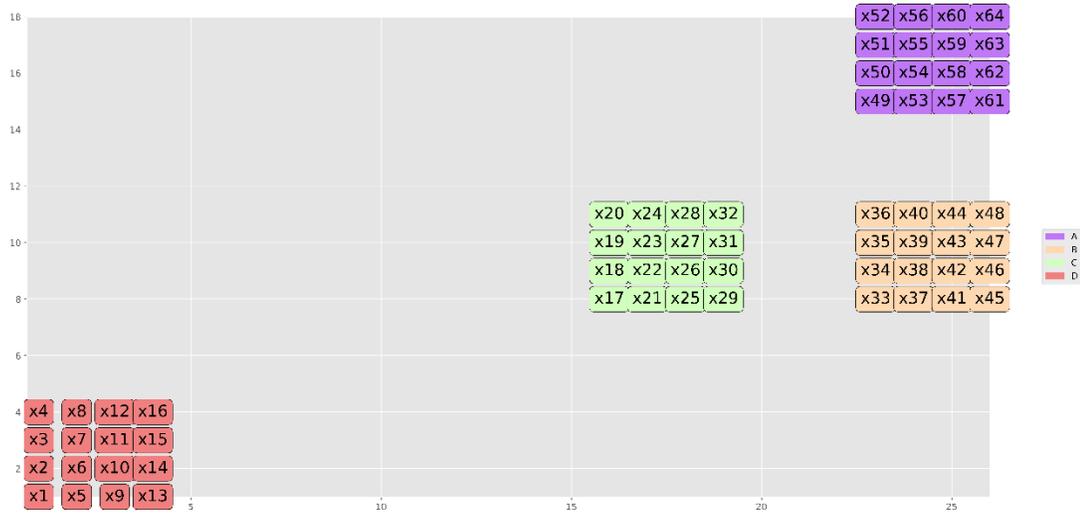

Figure 03: Dataset 1 Classes

As no assignment example is used the K-means++ algorithm is applied, and the following clusters and clusters centroids $C_k$ ($\theta_k; \rho_k$) are found:

- $C_1$ (17.5; 9.5) = $x17, x18, x19, x20, x21, x22, x23, x24, x25, x26, x27, x28, x29, x30, x31, x32$
- $C_2$ (2.5; 2.5) = $x1, x2, x3, x4, x5, x6, x7, x8, x9, x10, x11, x12, x13, x14, x15, x16$
- $C_3$ (24.5; 16.5) = $x49, x50, x51, x52, x53, x54, x55, x56, x57, x58, x59, x60, x61, x62, x63, x64$
- $C_4$ (24.5; 9.5) = $x33, x34, x35, x36, x37, x38, x39, x40, x41, x42, x43, x44, x45, x46, x47, x48$

Measuring each cluster centroid distance from the origin and ranking in decreasing order, we find that $C_3 > C_4 > C_1 > C_2$. Meaning that the alternatives in the cluster $C_3$ belongs to Class A, the alternatives in the cluster $C_4$ belongs to Class B, the alternatives in the cluster $C_1$ belongs to Class C and the alternatives in the cluster $C_4$ belongs to Class D.

In the next step, we set the following parameters: sample size is equal to 25%, and the number of models is equal to 1000, a total of 30 generations, the population size is equal to 15 chromosomes, and only 1 elite member. The operator parameters $\mu$ is equal to 2, and $\eta$ is equal to 1, and finally, the mutation rate is set to 5%. In Table 05 we show in detail one of the 1000 generated models.

Table 05: Dataset 1 generated model

| Parameters | Model-01 |
|---|---|
| Alternatives | x4, x7, x8, x13, x17, x19, x26, x28, x33, x35, x36, x38, x40, x50, x52, x64 |
| Weights | 0.53; 0.29 |
| Indifference | 1.13; 1.67 |
| Preference | 1.57; 6.73 |
| Veto | 3.45; 6.77 |
| Profile 1 | 1.24; 6.77 |
| Profile 2 | 19.59; 10.95 |
| Profile 3 | 20.83; 15.38 |
| Cutting Level | 0.86 |
| Accuracy | 100% |

This generated model has 16 alternatives that were randomly selected, the optimized weights favors $g_1$ as it is the criterion with the highest value, the indifference, preference, and veto thresholds respect the ELECTRE Tri-B constraints as the clipping operation in the GA algorithm guarantees that only feasible solutions are created. The same can be said concerning the reference profile limits, and this model has an accuracy of 100%, meaning that it could correctly separate all randomly selected alternatives in the classes defined by the ordered cluster.

The complete set of 1000 models has an average accuracy of 94%, with four models with an accuracy of 56%, three models with an accuracy of 63%, 13 models with an accuracy of 69%, 46 models with an accuracy of 75%, 99 models with an accuracy of 81%, 125 models with an accuracy of 88%, 189 models with an accuracy of 94% and 521 models with an accuracy of 100%. As most models have an accuracy value above 80%, the majority vote procedure is sufficient to assign an alternative to a class reliably. Although trimming bad models or tweaking the GA parameters to improve the models, is also a possibility and must be evaluated by the decision-maker considering he/she expertise about the problem.

Now taking into consideration all 1000 models, each alternative is voted; in other words, each model classifies the alternative in the four possible classes. The class with the majority votes is the one that the alternative must be assigned. Figure 04 shows the classification for each alternative and the decision boundaries (shaded regions). We can note that the voting procedure can be used as a non-linear classifier as it generates smooth curves to form the decision boundaries.

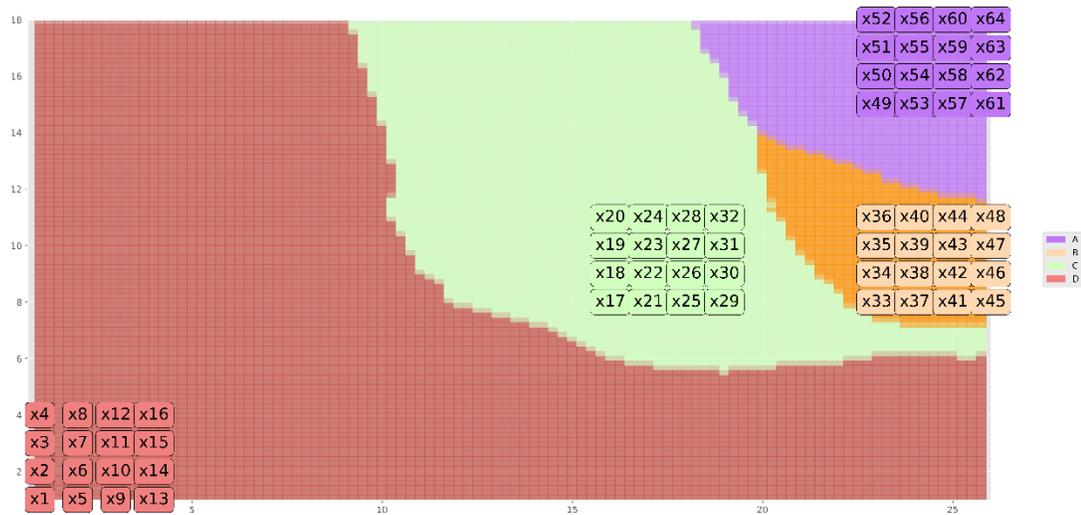

Figure 04: Dataset 1 non-linear decision boundaries

The merge procedure is done by taking the average of all values of each optimized model, and in this way, the required elicited parameters are found, as shown in Table 06.

Table 06: Dataset 1 elicitated parameters

| Elicitated Parameters | Values |
| --- | --- |
| Weights | 0.52; 0.49 |
| Indifference Threshold | 3.78; 2.70 |
| Preference Threshold | 5.83; 3.82 |
| Veto Threshold | 10.70; 5.68 |
| Profile 1 | 9.50; 8.20 |
| Profile 2 | 17.12; 10.68 |
| Profile 3 | 19.87; 13.45 |
| Cutting Level | 0.76 |
| Accuracy | 90.63% |

The elicitated parameters reduce the model to a standard ELECTRE Tri-B model, which can be understood as a linear classifier as the decision boundaries are defined as straight lines dividing each class. The accuracy is 90.63%, and trivially, for the same dataset, it can be inferred that the accuracy level of the merged procedure is equal or less the accuracy level of the voting procedure. Figure 05 shows the linear decision boundaries of the merge procedure.

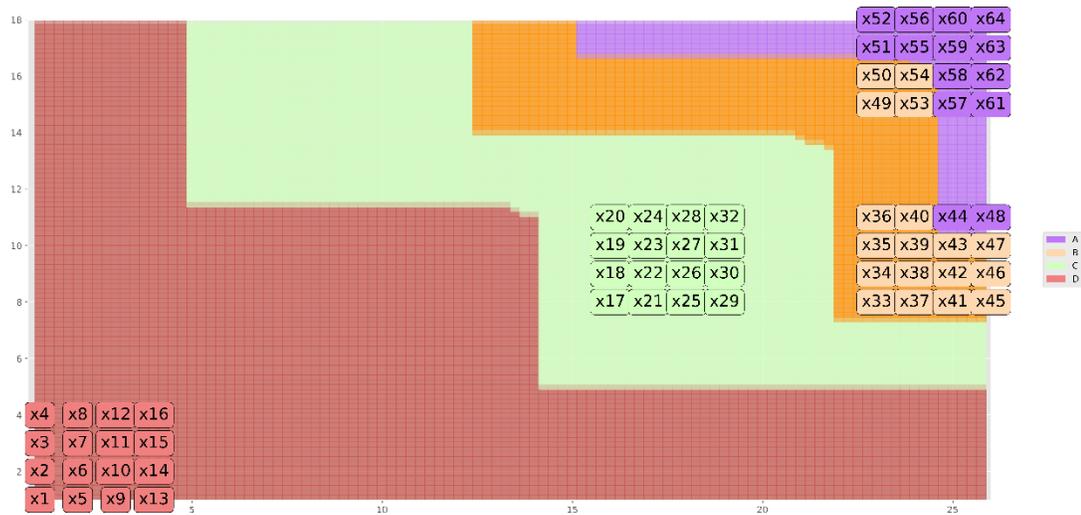

Figure 05: Dataset 1 linear decision boundaries

The second example is taken from Sobrie et al. (2019) that used The ESL (Employee Selection) dataset that is available at https://www.openml.org/d/1027, and it is has a total of 488 alternatives and 4 criteria of the maximization type ($g_1, g_2, g_3$ and $g_4$). According to Sobrie et al. (2019), each alternative corresponds to the profile of an applicant for a job and psychologists evaluated the applicants based on psychometric tests and interviews, to demonstrate their method the authors have normalized all 4 criteria. Furthermore, the dataset, as suggested by the authors, had their evaluation score binarized, separating the alternatives in two classes: "suitable" (Class A) and "not suitable" (Class B), where Class A > Class B. This version of the dataset is available at https://github.com/oso/pymcda/blob/master/datasets/esl.csv.

The authors have separated the dataset in two halves, 50% as a training set and 50% as a test set, and we did the same to compare the results obtained by their algorithm and the results obtained by our algorithm the ELECTRE Tree.

To visualize the dataset, we have projected it in a feature space with two dimensions, using the TSVD (Truncated Singular Value Decomposition) technique. In Figure 06, the left plot represents the training set, and the right plot represents the test set.

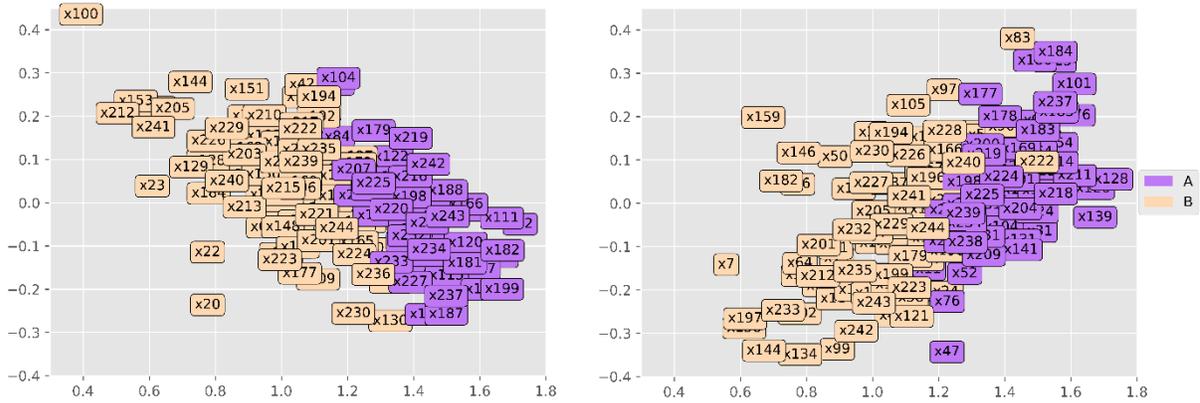

Figure 06: Dataset 2 training and test sets

The binarized evaluation score will serve as an assignment example, and we must elicitate the weights, lambda cutting level, veto threshold and the reference profiles for the classes A and B. The indifference and preference thresholds are equal zero for $g_1, g_2, g_3$ and $g_4$, and the pessimistic rule was used. We set the following parameters: sample size is equal to 10%, and the number of models is equal to 1000, a total of 250 generations, the population size is equal to 15 chromosomes, and only 1 elite member. The operator parameters $\mu$ is equal to 2, and $\eta$ is equal to 1, and finally, the mutation rate is set to 5%. The elicitated parameters are shown in Table 07.

Table 07: Dataset 2 elicitated parameters

| Parameters | Values |
| --- | --- |
| Weights | 0.49; 0.51; 0.54; 0.51 |
| Veto Threshold | 0.34; 0.33; 0.24; 0.21 |
| Profile1 | 0.56; 0.52; 0.59; 0.60 |
| Cutting Level | 0.71 |

Table 08 summarizes the results obtained by Sobrie et al. (2019), and the two versions of our algorithm. Concerning the training set, our voting procedure has the highest accuracy score, 93.03%, although it can be considered as a technical tie with Sobrie et al. (2019) model. Concerning the test set (generalization), both versions performed better; however, the model with elicitated parameters has the highest accuracy score with a result of 91.39%. This phenomenon occurs because the test set is different from the training set different. In this case, the merge procedure can have a better performance than the voting procedure as depending on the test set data structure (linear or non-linear); it may generalize better.

Table 08: Dataset 2 comparison

| | Sobrie et al. (2019) | ELECTRE Tree (Voting) | ELECTRE Tree (Merge) |
| --- | --- | --- | --- |
| Trainning set | 93.00% | **93.03%** | 91.80% |
| Test set | 88.11% | 90.16% | **91.39%** |

In Figure 07, to further demonstrate the non-linear characteristics of the ELECTRE Tree voting procedure, again using the TSVD technique, we have projected the decision boundary in a feature space with two dimensions. The plot in the left represents the voting procedure decision boundary, and the plot in the right represents the merged procedure decision boundary. The plot in the left is curvilinear with smooth edges in contrast with the plot in the right, which has sharp edges.

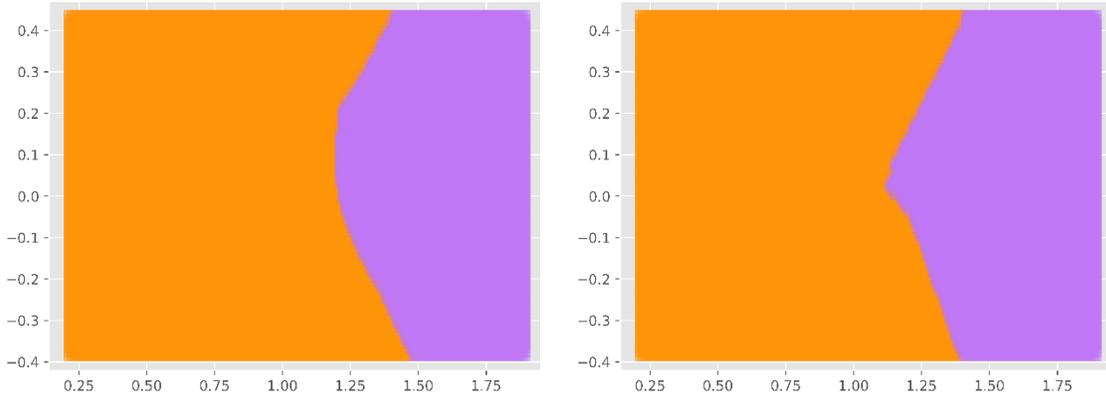

Figure 07: Dataset 2 decision boundaries

The third example is taken from Do Carvalhal Monteiro et al. (2020), the authors have collected the criteria from two different sources, World Development Indicators and UNICEF Program for Water Supply, Sanitation, and Hygiene. Both datasets were merged, resulting in eight variables (criteria) and 182 countries (alternatives) comprising the period from 2010 to 2015. The minimization criteria were transformed to maximization criteria by applying a reversed scale min-max normalization. The maximization criteria were normalized, applying a standard min-max normalization.

- *Neonatal Mortality Rate* ($g_1$ - minimization criterion): It is the number of neonates dying before reaching 28 days of age, per 1000 live births for a given year.
- *Under Five Mortality Rate* ($g_2$ - minimization criterion): It is the probability per 1000 that a newborn baby will die before reaching age five.
- *Immunization* ($g_3$ - Maximization criterion): It measures the percentage of children ages 12 - 23 months who received DPT (Diphtheria, Pertussis, and Tetanus) vaccinations.
- *Water* ($g_4$ - maximization criterion): It measures the percentage of individuals that have access to drinking water from an improved source, provided collection time is not more than 30 minutes for a round trip, including queuing.
- *Sanitation* ($g_5$ - maximization criterion): It measures the percentage of the use of improved facilities that are not shared with other households.
- *Pre-Primary Enroll* ($g_6$ - maximization criterion): It measures the gross enrollment of pre-primary education, which refers to programs designed primarily to introduce very young children to a school-type environment and to provide a bridge between home and school.
- *Primary Enroll* ($g_7$ - maximization criterion): It measures the gross enrollment of primary education which provides children with essential reading, writing, and mathematics skills along with an elementary understanding of such subjects as history, geography, natural science, social science, art, and music.

- *Primary Completion* ($g_8$ - maximization criterion): It is the number of new entrants (enrollments minus repeaters) in the last grade of primary education, regardless of age, divided by the population at the entrance age for the last grade of primary education. Data limitations preclude adjusting for students who drop out during the final year of primary education.

Our algorithm could be used to separate the countries in four classes (A = Very High, B = High, C = Low, D = Very Low) to analyze the general performance of countries about the child's well-being indicators.

We have selected the following parameters: sample size is equal to 10%, and the number of models is equal to 500, a total of 30 generations, the population size is equal to 15 chromosomes, and only 1 elite member. The operator parameters $\mu$ is equal to 2, and $\eta$ is equal to 1, and finally, the mutation rate is set to 5%.

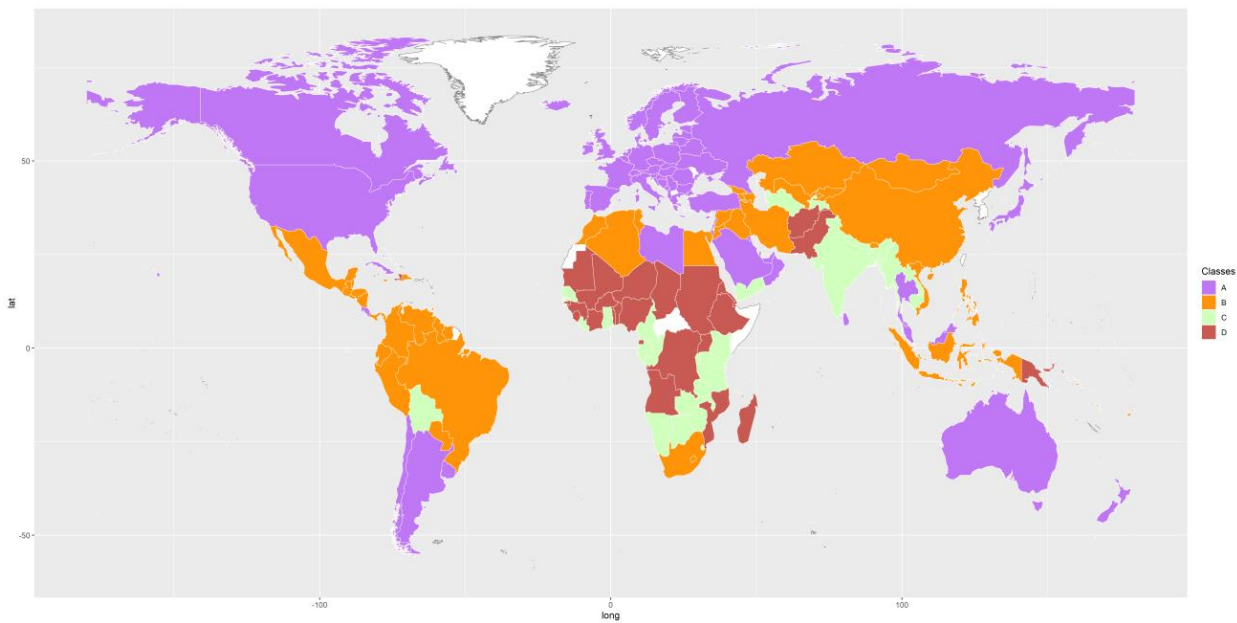

Figure 08: Dataset 3 Classes

Figure 08 indicates the results of the voting procedure; countries in Class D (Afghanistan, Angola, Haiti, Togo, etc.) demonstrate a weak effort to invest human capital during the individual life course. On the other hand, countries in Class A (Canada, Chile, Denmark, Finland, France, etc.) seem to see children as the future workforce and decision-makers and invest in human capital during childhood, seeking progress and development.

Finally, an implementation, in python 3.6, of the proposed algorithm was developed, and all examples are also available. It can be accessed at https://github.com/Valdecy/ELECTRE-Tree.

## 5 - Conclusion

Our proposed algorithm can be used in two versions, the first one a voting procedure can classify alternatives without the direct interference of the decision-maker, and it is highly indicated for problems with a non-linear data structure. Still, it also shows satisfactory results for problems with a linear data structure. The merge procedure is meant only for problems with linear data structure, and it has an advantage over the former procedure as it elicitates the parameters desired by the decision-maker. The decision-maker can accept the generated values or take them as a starting point to input their expertise about the problem and calibrate it at will.

Three examples were given, the first one was a toy example, used to clarify the steps of our algorithm and demonstrate transparently how the decision boundaries are formed. The second example was a real case adapted for the ELECTRE Tri-B environment; in other words, a multiple criteria problem with monotone classes. Our algorithm performed well when compared to the results of the literature showing that a decision-maker could confidently use the merge or voting procedure results to solve a problem and adding new alternatives if needed. The final example shows the use of our procedure as a stand-alone multicriteria tool, aiding the decision-maker in a multicriteria problem.

The generate models usually can obtain high accuracy values, that can be improved by tweaking the GA parameters, or the decision-maker can just trim the undesired models. Yet if the number of bad models is low, both voting and merge procedures can be done without invalidating the final solution or requiring to give different weights for each model. Generating a broad set of models (between 200 and 1000) usually is indicated because of the sample size, as is it is randomly sampled with replacement, doing it many times guarantees that the ensemble of models can represent all data structure. For future studies, other types of ordered clusters algorithms can be investigated, and also the creation of specialized heuristics to be used in the optimization phase.

**REFERENCES**

ARTHUR, D.; VASSILVITSKII, S. (2007). K-means++: The Advantages of Careful Seeding. Proceedings of the Eighteenth Annual ACM-SIAM Symposium on Discrete Algorithms (SODA 2007). Society for Industrial and Applied Mathematics Philadelphia, PA, USA. pp. 1027–1035. doi: 10.1145/1283383.1283494

BURKE, E.K.; VARLEY, D.B. (1997). A Genetic Algorithms Tutorial Tool for Numerical Function Optimisation. Proceedings of the 2nd Conference on Integrating Technology into Computer Science Education (ITiCSE). doi: 10.1145/268819.268830

CAILLOUX, O., MEYER, P., MOUSSEAU, V. (2012). Eliciting ELECTRE TRI Category Limits for a Group of Decision Makers. European Journal of Operational Research. Vol. 223, pp. 133 - 140. doi: 10.1016/j.ejor.2012.05.032

DEB, K.; AGRAWAL, R B. (1995). Simulated Binary Crossover for Continuous Search Space. Complex Systems. Vol. 9, Issue 2, pp. 115 – 148, doi: 10.1.1.26.8485

DIAS L.C.; MOUSSEAU, V. (2002). Inferring ELECTRE's Veto Related Parameters from Outranking examples. Research Report No 5/2002, INESC Coimbra.

DO CARVALHAL MONTEIRO, R.L.; PEREIRA, V.; COSTA, H.G. (2020) Dependence Analysis Between Childhood Social Indicators and Human Development Index Through Canonical Correlation Analysis. Child Indicators Research, Vol. 13, pp. 337 - 362. doi: 10.1007/s12187-019-09715-6

DOUMPOS, M.; MARINAKIS, Y.; MARINAKI, M.; ZOPOUNIDIS, C. (2009). An Evolutionary Approach to Construction of Outranking Models for Multicriteria Classification: The Case of the ELECTRE TRI Method. European Journal of Operational Research. Vol. 199, pp. 496 - 505. doi:10.1016/j.ejor.2008.11.035

FERNÁNDEZ, E.; FIGUEIRA, J. R., NAVARRO, J. (2019). An Indirect Elicitation Method for the Parameters of the ELECTRE TRI-nB Model Using Genetic Algorithms. Applied Soft Computing. Vol. 77, April 2019, pp. 723 - 733. doi: 10.1016/j.asoc.2019.01.050

GOVINDAN, K.; JEPSEN, M. B. (2016). ELECTRE: A Comprehensive Literature Review on Methodologies and Applications. European Journal of Operational Research. Vol. 250, pp. 1-29. doi: 10.1016/j.ejor.2015.07.019

HO, T. K. (1995). Random Decision Forests. Proceedings of the 3rd International Conference on Document Analysis and Recognition, Montreal, QC, 14 - 16 August 1995. pp. 278 - 282. doi: 10.1109/ICDAR.1995.598994

HOLLAND, J.H: (1973). Genetic Algorithms and the Optimal Allocation of Trials. Society for Industrial and Applied Mathematics Journal on Computing. Vol. 2, Issue 2, pp. 88 – 105. doi: 10.1137/0202009

JERÔNIMO, T.B., MEDEIROS, D. (2014). Measuring Quality Service: The Use of a SERVPERF Scale as an Input for ELECTRE TRI Multicriteria Model. International Journal of Quality & Reliability Management Vol. 31, Issue 6, pp. 652 - 664. doi 10.1108/IJQRM-06-2012-0095

KONAK, A.; COIT, D.W.; SMITH, A.E. (2006). Multi-objective Optimization Using Genetic Algorithms: A tutorial. Reliability Engineering and System Safety. Vol. 91, pp. 992 - 1007. doi: 10.1016/j.ress.2005.11.018

LEONE, R.; MINNETTI, V. (2013). The Estimation of the Parameters in Multi-Criteria Classification Problem: The Case of the ELECTRE Tri Method. Analysis and Modeling of Complex Data in Behavioral and Social Sciences. doi: 10.1007/978-3-319-06692-9_11.

LEROY, A.; MOUSSEAU, V.; PIRLOT, M. (2011). Learning the Parameters of a Multiple Criteria Sorting Method Based on a Majority Rule. Conference Paper. doi: 10.1007/978-3-642-24873-3_17

LUCASIUS, C.B.; KATEMAN, G. (1993). Understanding and Using Genetic Algorithms Part 1: Concepts, Properties and Context. Chemometrics and Intelligent Laboratory Systems. Vol. 19, pp. 1 - 33. doi: 10.1016/0169-7439(93)80079-W

MCQUEEN, J. (1967). Some Methods for classification and Analysis of Multivariate Observations. Proceedings of 5th Berkeley Symposium on Mathematical Statistics and Probability. Vol. 1, Statistics, pp. 281-297. University of California Press, Berkeley, Calif. Mathematical Reviews number: MR0214227. Available at: < https://projecteuclid.org/euclid.bsmsp/1200512992>

MENDOZA, G. A.; MARTINS, H. (2006). Multi-criteria Decision Analysis in Natural Resource Management: A Critical Review of Methods and New Modelling Paradigms. Forest Ecology and Management, Vol. 230, Issues 1-3, pp. 1-22. doi: 10.1016/j.foreco.2006.03.023.

MOUSSEAU, V.; FIGUEIRA, J.; NAUX, J.P.H. (2001). Using Assignment Examples to Infer Weights for ELECTRE TRI Method: Some Experimental Results. European Journal of Operational Research. Vol. 130, pp. 263 – 275. doi: 10.1016/S0377-2217(00)00041-2

MOUSSEAU, V.; SLOWINSKI, R. (1998). Inferring an ELECTRE TRI Model from Assignment Examples. Journal of Global Optimization. Vol. 12, pp. 157 - 174. doi: 10.1023/A:1008210427517

MOUSSEAU, V.; SLOWINSKI, R.; ZIELNIEWICZ, P. (2000). A User-oriented Implementation of the ELECTRE-TRI Method Integrating Preference Elicitation Support. Computers & Operations Research. Vol. 27, pp. 757 -777. doi: 10.1016/S0305-0548(99)00117-3

NGO THE, A.; MOUSSEAU. V. (2002). Using Assignment Examples to Infer Category Limits for the ELECTRE TRI Method. Journal of Multi-Criteria Decision Analysis. Vol. 11, pp. 29 – 43. doi: 10.1002/mcda.314

PEÑA, J.M., LOZANO, J.A., LARRAÑAGA, P. (1999). An Empirical Comparison of Four Initialization Methods for the K-Means Algorithm. Pattern Recognition Letters. Vol. 20, pp.1027 - 1040. doi: 10.1016/S0167-8655(99)00069-0

PIRLOT, M.(1996). General Local Search Methods. European Journal of Operational Research. Vol. 92, Issue 3, pp. 493 - 511. doi: 10.1016/0377-2217(96)00007-0

RAMEZANIAN, R. (2019). Estimation of the Profiles in Posteriori ELECTRE TRI: A Mathematical Programming Model. Computers & Industrial Engineering. Vol. 128, pp. 47 - 59. doi: .10.1016/j.cie.2018.12.034

RIPON K.S.N.; KWONG S.; MAN K.F. (2007). A Real-coding Jumping Gene Genetic Algorithm (RJGGA) for Multiobjective Optimization. Information Sciences. Vol. 177, Issue 2, pp. 632 - 654. doi: /10.1016/j.ins.2006.07.019

ROY, B.; BOUYSSOU, D. (1993). Aide Multicritère à la Décision: Méthodes et Cas. Economica, Paris, France.

ROY, B.; FIGUEIRA, J.R.; ALMEIDA-DIAS, J. (2014). Discriminating Thresholds as a Tool to Cope with Imperfect Knowledge in Multiple Criteria Decision Aiding: Theoretical Results and Practical Issues. Omega, Vol. 43, pp; 9 – 20. doi: 10.1016/j.omega.2013.05.003

SOBRIE, O.; MOUSSEAU, V.; PIRLOT, M. (2019). Learning Monotone Preferences Using a Majority Rule Sorting Model. International Transactions in Operational Research. Vol. 26, Issue 5, pp. 1786 – 1809. doi: 10.1111/itor.12512

WANG, J. J.; JING, Y. Y.; ZHANG, C. F.; ZHAO, J. H. (2009). Review on Multi-criteria Decision Analysis Aid in Sustainable Energy Decision-making. Renewable and Sustainable Energy Reviews. Vol. 13, Issue 9, pp. 2263 – 2278. doi: 10.1016/j.rser.2009.06.021

YU, W. (1992). ELECTRE TRI: Aspects Méthodologiques et Manuel d'Utilisation. Document du LAMSADE Nº 74, Université Paris-Dauphine, Paris, France.

ZHENG, J. (2012). Elicitation des Préférences pour des Modèles d'Agrégation basés sur des Points de Référence: Algorithmes et Procédures. École Centrale Paris. ⟨NNT: 2012ECAP0028⟩. Available at: ⟨ https://tel.archives-ouvertes.fr/tel-00740655 ⟩

ZHENG, J.; TAKOUGANG, S.A.M.; MOUSSEAU, V.; PIRLOT, M. (2014). Learning Criteria Weights of an Optimistic ELECTRE TRI Sorting Rule. Computers & Operations Research. Vol. 49, pp. 28 - 40. doi: 10.1016/j.cor.2014.03.012